\documentclass[a4paper, 10pt, conference]{ieeeconf}      
\pdfoutput=1
\IEEEoverridecommandlockouts                              

\usepackage{bm}
\usepackage[capitalize]{cleveref}
\usepackage{mathrsfs}
\usepackage{todonotes}
\usepackage{amsmath}
\usepackage[detect-all,per-mode=symbol]{siunitx}
\usepackage{amsfonts}
\let\svtodo\todo\renewcommand\todo[1]{\svtodo[inline]{#1}}
\usepackage[style=ieee]{biblatex}
\usepackage{booktabs}

\newcolumntype{P}[1]{>{\centering\arraybackslash}p{#1}}

\title{\LARGE \bf
Adaptive Stress Testing with Reward Augmentation\\ for Autonomous Vehicle Validation
}

\author{Anthony Corso,$^*$ Peter Du,$^*$ Katherine Driggs-Campbell, and Mykel J. Kochenderfer
\thanks{*These authors contributed equally.}
\thanks{A. Corso and M. J. Kochenderfer are with the Aeronautics and Astronautics Department, Stanford University. e-mail: \{acorso,mykel\}@stanford.edu}%
\thanks{P. Du and K. Driggs-Campbell are with the Electrical and Computer Engineering Department, University of Illinois at Urbana-Champaign. e-mail: \{peterdu2,krdc\}@illinois.edu}%
}

\begin{document}

\maketitle
\thispagestyle{empty}
\pagestyle{empty}

\begin{abstract}
Determining possible failure scenarios is a critical step in the evaluation of autonomous vehicle systems. Real world vehicle testing is commonly employed for autonomous vehicle validation, but the costs and time requirements are high. Consequently, simulation driven methods such as Adaptive Stress Testing (AST) have been proposed to aid in validation. AST formulates the problem of finding the most likely failure scenarios as a Markov decision process, which can be solved using  reinforcement learning. In practice, AST tends to find scenarios where failure is unavoidable and tends to repeatedly discover the same types of failures of a system. This work addresses these issues by encoding domain relevant information into the search procedure. With this modification, the AST method discovers a larger and more expressive subset of the failure space when compared to the original AST formulation. We show that our approach is able to identify useful failure scenarios of an autonomous vehicle policy.
\end{abstract}

\section{INTRODUCTION}

Validation of autonomous vehicles (AVs) is a critical and challenging task that needs to be addressed to ensure the safety of human drivers and vulnerable participants such as pedestrians. AV validation remains difficult due to the wide variety of driving scenarios that need to be handled by the AV. While real world vehicle testing is commonly used for validation, the costs and time requirements are extremely high~\cite{Koopman2016challenges,Kalra2014,Junietz2018}. Even with simulation, the space of driving scenarios is too expansive for naive sampling techniques to adequately cover. Various adaptive sampling approaches have been explored~\cite{MULLINS2018197,OKelly2018ScalableEA,8500421,7571159}.

One adaptive simulation-based technique for autonomous system validation is Adaptive Stress Testing (AST)~\cite{lee2015adaptive}. AST involves finding the most likely failures of a system by formulating the search through the space of scenarios as a Markov decision process (MDP) and applying reinforcement learning (RL) methods to find a solution. The reward function of the MDP depends on whether the autonomous system experiences a failure (i.e., a vehicle collision), and the likelihood of state transitions. This reward formulation has two issues:
\begin{enumerate}
\item AST finds scenarios where failure is unavoidable (e.g. a pedestrian causing a collision with a stopped AV).
\item AST tends to repeatedly discover failures that are similar to ones that it already found.
\end{enumerate}
These outcomes are unhelpful for engineers and policy makers who want to find a variety of different failure scenarios where the autonomous vehicle should have behaved differently. This work seeks to address these issues by modifying the reward function. We incorporate heuristics related to autonomous driving to guide the search algorithm toward the most relevant failures.

We employ two approaches to help find the most informative failures of an autonomous driving scenario using AST. To address the first issue, we incorporate the Responsibility-Sensitive Safety (RSS) policy into the reward function~\cite{shalev2017formal}. RSS is a formal, mathematically interpretable model to characterize autonomous vehicle safety. By incorporating RSS into the reward function, we help guide AST to find scenarios where the vehicle behaved improperly prior to a collision. To address the second issue, we incorporate a dissimilarity metric that encourages the discovery of failures that are highly distinct from one another. These reward modifications allow AST to produce a wider range of failures that are of greater interest when compared with results obtained using the existing AST reward function.

 We present the following contributions:
\begin{enumerate}
	\item We propose a framework to extend AST for autonomous vehicle validation through reward augmentation to better identify relevant failures. 
	\item We present two augmentation techniques that can be used with the AST method. The first uses the RSS framework to identify specific instances of improper AV behavior. The second incorporates a dissimilarity metric to find multiple, unique modes of failure involving the AV. 
	\item We apply AST with an augmented reward on a scenario involving an autonomous vehicle and pedestrians and show the effectiveness of our methods in finding relevant failure cases when compared with an existing AST setup. 
\end{enumerate}

This paper is organized as follows: \cref{sec:background} provides an overview of the AST method and how it is applied to an autonomous vehicle scenario, as well as the RSS rule set. \Cref{sec:methods} presents the reward augmentation methods for RSS and the dissimilarity metric. \cref{sec:experiments} describes several experiments and their results. Lastly, \cref{sec:conclusions} provides concluding remarks. 

\section{BACKGROUND}
\label{sec:background}
This section provides the necessary background for understanding how AST is used with an autonomous driving scenario, as well as the relevant rules of RSS.

\subsection{Adaptive Stress Testing}
\label{ast}
Adaptive stress testing is a validation technique that frames the problem of validating an autonomous policy (referred to as system-under-test or SUT) interacting with an environment as a Markov decision process. Various reinforcement learning algorithms can then be applied to find the most probable failure modes of the SUT. AST has successfully been applied to aircraft collision avoidance systems \cite{lee2015adaptive, lee2018differential}
and autonomous driving policies \cite{koren2018adaptive}. The AST method is detailed below with an example of an autonomous driving policy being validated for collision avoidance. 

AST requires a simulator $\mathscr{S}$ of the SUT interacting with an environment and a set of possible actions $\mathcal{A}$. For example, the SUT could be an autonomous driving policy controlling a car that is approaching a busy intersection. The action space of the environment could include, for example, the pedestrians' accelerations, the friction coefficient of the road, the trajectories of other cars, and the amount of sensor noise in the perception system. The user of AST defines a set of critical states $E$ that are failures of the autonomous policy. In the driving example, $E$ may contain all states of the simulator where the AV has approached too close to, or collided with, another agent. 

The simulator can be a black-box system that does not make its internal operation and state available to AST, but it must expose the following interface:
\begin{itemize}
    \item \verb|Initialize|$(\mathscr{S})$: Resets simulator to a starting state.
    \item \verb|Step|$(\mathscr{S}, a)$: Steps the simulation forward in time by taking action $a$. Returns a flag indicating if the new state of the simulator is in $E$.
    \item \verb|IsTerminal|$(\mathscr{S})$: Checks whether the current state of the system is in the critical set $E$.
\end{itemize}
AST operates using the simulator $\mathscr{S}$, a reward function, and an RL solver. The solver generates environment actions that are used to update $\mathscr{S}$. Once in its new state, $\mathscr{S}$ checks if a critical state has been reached or if the simulation has reached its maximum duration, then passes that information to the reward function. The reward function then generates a reward and passes it back to the solver.

\begin{figure}
    \centering
    \includegraphics[width=0.7\columnwidth]{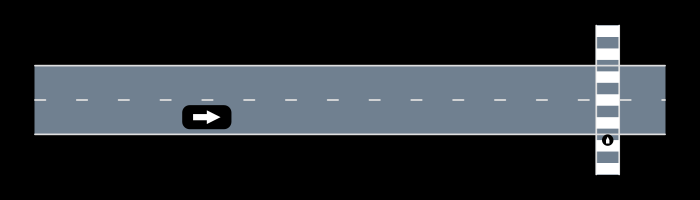}
    \caption{Autonomous vehicle crosswalk scenario~\cite{koren2018adaptive}.}
    \label{fig:av_scenario}
    \vskip -1cm
\end{figure}

The AV scenario under study is based on the work of \citeauthor{koren2018adaptive} and is shown in 
\cref{fig:av_scenario}, where an autonomous vehicle approaches a crosswalk where one or more pedestrians is attempting to cross. The SUT is the autonomous vehicle controlled by the Intelligent Driver Model (IDM) ---  an algorithm that keeps the vehicle in its lane, following the traffic ahead while maintaining a safe distance \cite{Treiber2000congested}. The goal of the vehicle is to identify the pedestrians crossing the road and apply adequate braking to avoid coming too close. The state of each agent is represented by a 4-tuple:
$$
s_{\rm agent} = (v_x, v_y, x, y)
$$
\noindent where $(v_x, v_y)$ represents the agent's velocity and $(x,y)$ represents the agent's position.

AST has control over pedestrian acceleration and the noise in the AV's sensors. In particular, the action space associated with the $i$th pedestrian is represented by a 6-tuple:

$$
a^{(i)} = (a_x^{(i)}, a_y^{(i)}, n_{v_x}^{(i)}, n_{v_y}^{(i)}, n_{x}^{(i)}, n_{y}^{(i)})
$$
\noindent where, for agent $i$, $(a_x^{(i)}, a_y^{(i)})$ is the agent's acceleration, $(n_{v_x}^{(i)}, n_{v_y}^{(i)})$ is the sensor noise corresponding to the agent's velocity, and $(n_{x}^{(i)}, n_{y}^{(i)})$ is the sensor noise corresponding to the agent's position. The noise components are processed at each time step and added to the vehicle's observation of each pedestrian's position and velocity. Actions are sampled by the AST solver from a uniform distribution where the acceleration values are between [-1,1] and noise values are between [0,1].

The reward function is defined as follows:
\begin{equation}
\label{eq:ast_reward}
R\left(s\right) = \left\{
        \begin{array}{ll}
            0 &  s \in E \\[7pt]
            -\alpha - \beta \mathcal{D}(r_v, r_p) &  s \notin E, t\geq T \\[7pt]
            -\mathcal{M}(a \mid s) &  s \notin E, t < T
        \end{array}
    \right.
\end{equation}
where $\mathcal{D}(r_v, r_p)$ is the Euclidean distance between the autonomous vehicle and the pedestrian and $\mathcal{M}(a \mid s) $ is the Mahalanobis distance of the action $a$ given the current state $s$, which is related to the $\log$ probability of that action ~\cite{P.C.}. The constants $\alpha$ and $\beta$ are set to \num{10000} and \num{1000}, respectively, to penalize the algorithm for not finding a collision. Solvers used in our application include Monte Carlo Tree Search (MCTS)~\cite{Browne2012} and Trust Region Policy Optimization (TRPO) ~\cite{schulman2015trust} because both have been shown to successfully find failures when combined with AST~\cite{lee2015adaptive,koren2018adaptive}.

\subsection{Responsibility-Sensitive Safety}
Responsibility-Sensitive Safety (RSS) is a set of driving rules motivated by common-sense driving practices that, when all agents on the road follow them, produce a driving utopia where no collisions will occur~\cite{shalev2017formal}. It is unrealistic to assume that all drivers will follow these rules exactly at all times, but RSS can help us formalize the responsibility of each party involved in an accident. RSS designates an agent as responsible for an accident if it did not follow the rules governing a proper response when the driving situation became unsafe. A response is determined to be proper if it is permitted by the RSS rules. 

For the purposes of this work, a subset of RSS rules are selected, specifically those concerned with longitudinal and lateral collision avoidance. Consider two agents on the road $c_1$ and $c_2$ with longitudinal velocities $v_1$ and $v_2$. If the response time, maximum possible acceleration, and maximum possible braking of the agents is known, then the minimum safe distance between them can be derived from simple kinematic relationships~\cite{shalev2017formal}. If the agents are moving in the same direction, then the minimum safe longitudinal distance between them is given by:
\begin{equation}
    d = \left[ v_1 \rho + \frac{1}{2} a_{\rm max}^{\rm acc} \rho^2 + \frac{\left(v_1 + \rho a_{\rm max}^{\rm acc}\right)^2}{2 a_{\rm min}^{\rm brk}} - \frac{v_2^2}{2 a_{\rm max}^{\rm brk}} \right]_+
\end{equation}
where $[x]_+ = \max(x,0)$, $\rho$ is the agents' response time, $a_{\rm max}^{\rm acc}$ is the maximum allowed longitudinal acceleration of either vehicle, $a_{\rm min}^{\rm brk}$ is the minimum longitudinal braking required to avoid a collision, and $a_{\rm max}^{\rm brk}$ is the maximum longitudinal braking possible by a vehicle. If $c_1$ and $c_2$ are moving toward one another (with $c_2$ having a negative velocity), then the safe longitudinal distance is given by:
\begin{equation}
d = \frac{v_1 + v_{1,\rho}}{2}\rho + \frac{v_{1,\rho}^2}{2 a_{\rm min}^{\rm brk,1}} + \frac{\lvert v_2 \rvert + v_{2,\rho}}{2}\rho  + \frac{v_{2,\rho}^2}{2 a_{\rm min}^{\rm brk, 2}}
\end{equation}
where $v_{1,\rho} = v_1  + \rho a_{\rm max}^{\rm acc}$ and $v_{2,\rho} = \lvert v_2\rvert + \rho a_{\rm max}^{\rm acc}$.

A driving situation is considered \textit{longitudinally dangerous} if the distance between $c_1$ and $c_2$ is less than the safe distance $d$. Let $t_d^{\rm long}$ be the time at which a situation first becomes longitudinally dangerous, then a \textit{longitudinal proper response} is one that adheres to the following rules:

\begin{enumerate}
    \item In the pre-response interval [$t_d^{\rm long}$, $t_d^{\rm long} + \rho$], cars may accelerate toward each other by no more than $a_{\rm max}^{\rm acc}$.
    \item During the response time $(t > t_d^{\rm long} + \rho)$, $c_1$ must brake at least as hard as $-a_{\rm min}^{\rm brk}$ until reaching a safe situation and then can have any non-positive acceleration.
    \item During the response time, if $c_2$ has positive velocity, then it cannot brake harder than $-a_{\rm max}^{\rm brk}$. If $c_2$ has negative velocity then it must brake at least as hard as $-a_{\rm min}^{\rm brk}$. Once a safe situation is reached, any non-negative acceleration is allowed. 
\end{enumerate}

A similar analysis can be done in the lateral direction. Assuming that $c_1$ is to the left of $c_2$, then the minimum safe lateral distance is:
\begin{equation}
\begin{split}
    d =
\left[\frac{v_1 + v_{1,\rho}}{2} \rho + \frac{v_{1,\rho}^2}{2 a_{\rm min}^{\rm brk}} -  \frac{v_2 + v_{2,\rho}}{2}\rho + \frac{v_{2,\rho}^2}{2 a_{\rm min}^{\rm brk}} \right]_+
    \end{split}
\end{equation}
where velocities and accelerations are now defined for the lateral direction. A situation is considered \textit{laterally dangerous} if the lateral distance between two agents is less than the safe lateral distance. Let $t_d^{\rm lat}$ be the time at which a situation becomes laterally dangerous, then a \textit{lateral proper response} requires adherence to the following rules:

\begin{enumerate}
    \item In the pre-response interval [$t_d^{\rm lat}$, $t_d^{\rm lat} + \rho$], the cars may accelerate toward each other by no more than $a_{\rm max}^{\rm acc}$.
    \item During the response time $(t > t_d^{\rm long} + \rho)$, $c_1$ and $c_2$ must each brake laterally at least as hard as $a_{\rm min}^{\rm brake}$ to bring their lateral velocity toward 0. After reaching a lateral velocity of 0, $c_1$ can have any non-positive lateral acceleration and $c_2$ any non-negative acceleration.
\end{enumerate}

Finally, we can combine these rules into a single rule for basic obstacle avoidance. Let $t_d$ be the first time that a situation is both laterally and longitudinally dangerous. A \textit{proper response} for agents $c_1$ and $c_2$ is as follows:
\begin{enumerate}
    \item If $t_d = t_d^{\rm long}$, then $c_1$ and $c_2$ employ a longitudinal proper response.
    \item If $t_d = t_d^{\rm lat}$, then $c_1$ and $c_2$ employ a lateral proper response.
\end{enumerate}

\begin{figure}[ht]
    \centering
    \includegraphics[width=\columnwidth]{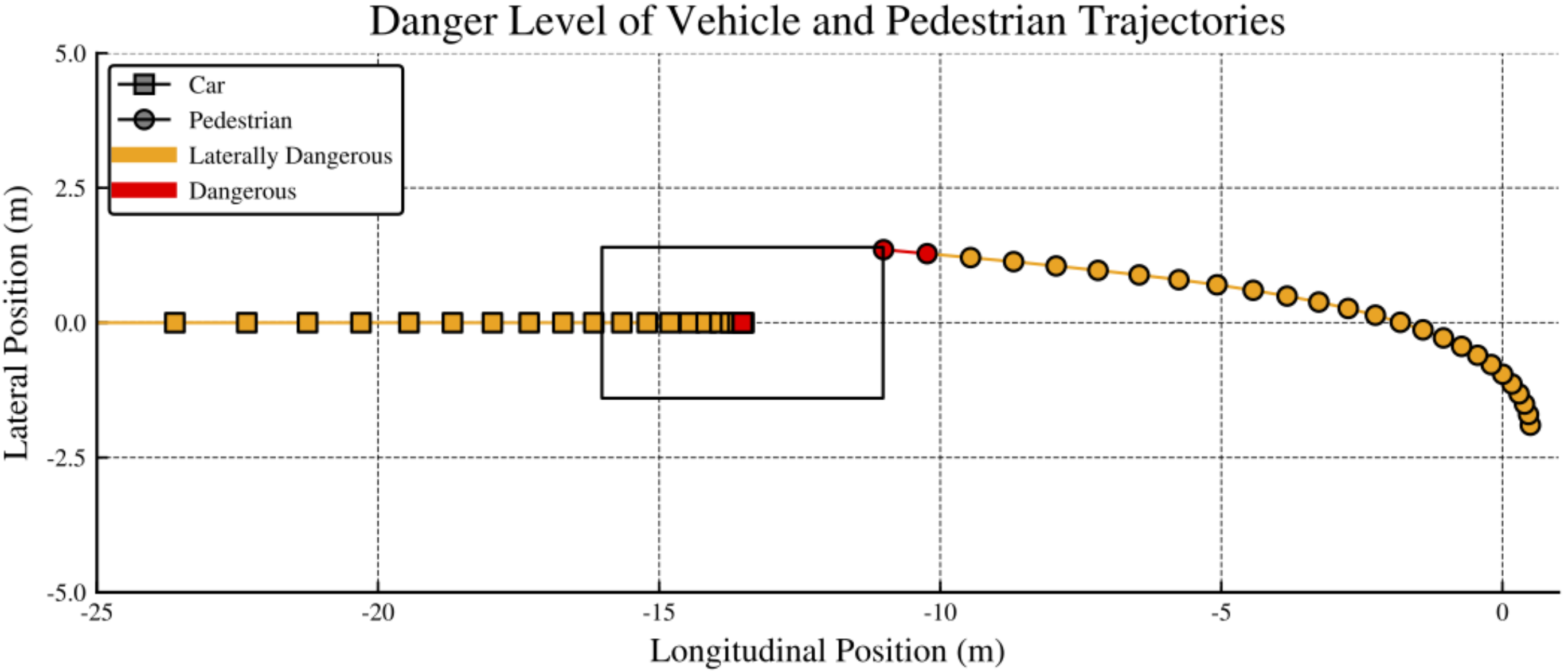}
    \caption{Example trajectory with the danger classification at each timestep.}
    \label{fig:example_danger}
\end{figure}

\begin{figure}[ht]
    \centering
    \includegraphics[width=\columnwidth]{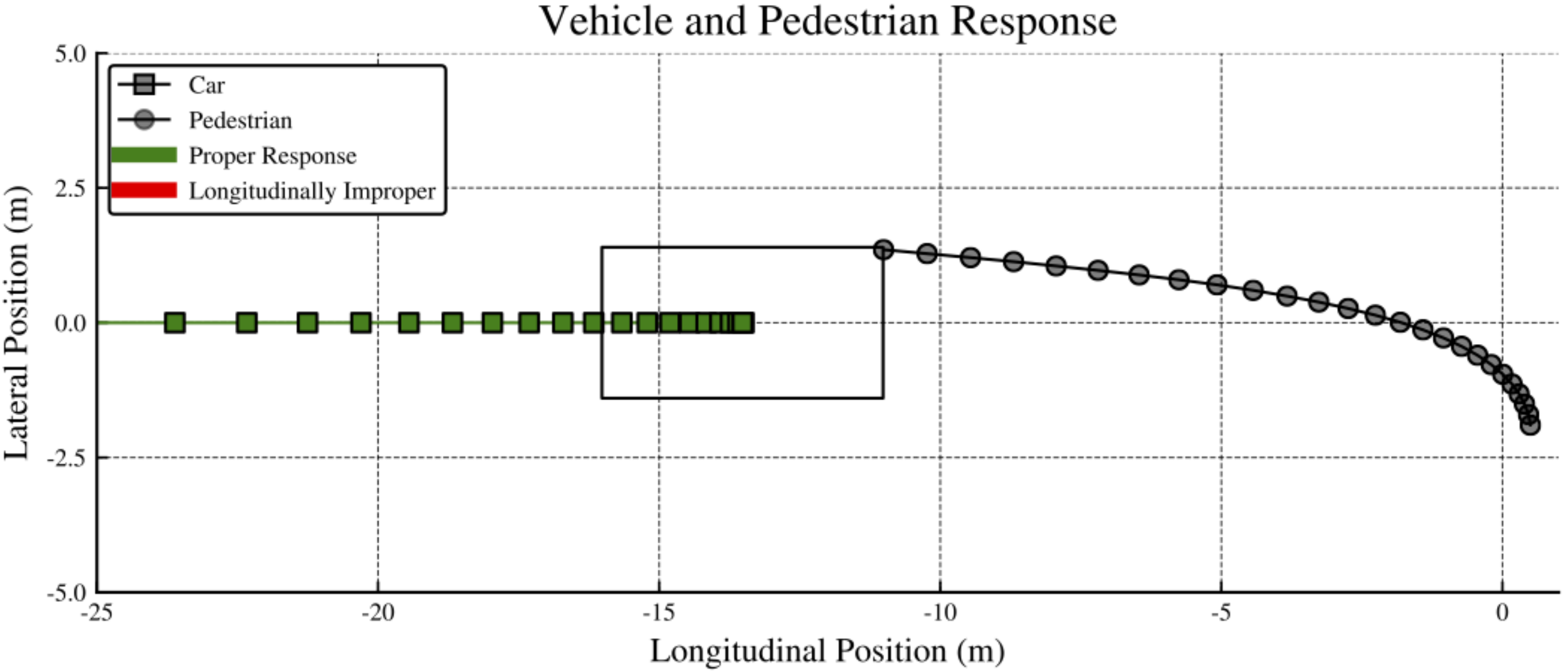}
    \caption{Example trajectory with autonomous vehicle response classification at each timestep.}
    \label{fig:example_response}
\end{figure}

Using this formulation, we can classify the actions of an agent as either proper or improper at any timestep in the simulation. \cref{fig:example_danger,fig:example_response} show an example interaction between a vehicle and a pedestrian that results in a collision. These simulations are a re-creation of the scenario shown in \cref{fig:av_scenario}. In these plots, the vehicle moves from left to right and the pedestrian moves from right to left. The scenario ends in a collision where the pedestrian collides with the front corner of the car. \Cref{fig:example_danger} shows timesteps that are colored based on category of danger (longitudinally dangerous, laterally dangerous, or both), while \cref{fig:example_response} shows colors indicating if the AV is behaving properly. The driving situation starts out laterally dangerous and only becomes longitudinally dangerous as the pedestrian gets close to the car. The car applied strong braking to miss the pedestrian. Its behavior is classified as proper with respect to the vehicle at every timestep, despite the trajectories ending in a collision. 

\section{METHODS}
\label{sec:methods}

This section outlines two reward augmentation approaches used on AST. The first uses RSS classification of proper and improper behavior and the second uses a dissimilarity metric of state trajectories that end in a failure. 

\subsection{RSS Rewards}

We define $f_{\rm imp}$ as the fraction of timesteps considered improper by RSS for the state trajectory of the autonomous vehicle. The set of critical events $E$ is altered to only include events that result from trajectories where the AV has behaved improperly. Let $\tau$ be a sequence of actions applied to the simulator and define:
\begin{equation}
E_{\rm RSS} = \left\{\tau \mid \tau \in E \ \rm{and} \ f_{\rm imp}(\tau) > f_{\rm crit} \right\}
\end{equation}
where $0 \leq f_{\rm crit} < 1$ is a tunable parameter that governs how many timesteps the vehicle has behaved improperly before it is considered at fault. The RSS reward function is then given by:
\begin{equation}
\label{eq:rss_reward}
R_{\rm RSS}\left(s\right) = \left\{
        \begin{array}{ll}
            0 &  s \in E_{\rm RSS} \\[7pt]
            -\alpha - \beta f_{\rm imp}(\tau) &  s \notin E_{\rm RSS}, t\geq T \\[7pt]
            -\mathcal{M}(a \mid s) &  s \notin E_{\rm RSS}, t < T
        \end{array}
    \right.
\end{equation}
where we replaced the Euclidean distance metric $\mathcal{D}$ with a metric that depends upon the fraction of timesteps that are improper in that trajectory. We are making the assumption that trajectories that have a higher fraction of timesteps where the autonomous vehicle behaved improperly are more likely to end in a collision. For RSS to classify an action as improper, a collision must already be imminent, and the vehicle is not appropriately resolving the situation.

\subsection{Trajectory Dissimilarity (TD) Rewards}

Suppose $\tau_1$ and $\tau_2$ are the spatial trajectories of two agents (possibly of different lengths) represented by a sequence of points in $\mathbb{R}^2$ such that $\tau_1 = \{p_1^1, p_1^2, \ldots, p_1^i\}$ and $\tau_2 = \{p_2^1, p_2^2, \ldots, p_2^j\}$. We first normalize the lengths of both trajectories by dividing each trajectory into $n$ trajectory segments such that each segment contains a consecutive sub-sequence from either $\tau_1$ or $\tau_2$. For each segment, a \textit{representative} $c_x^i: i\in [1,n]$ is then calculated by computing the center of mass of the segment \cite{Liu2013}. The trajectories are now expressed as a sequence of these representative points such that $\tau_1' = \{c_1^1, c_1^2, \ldots, c_1^n\}$ and $\tau_2' = \{c_2^1, c_2^2, \ldots, c_2^n\}$. The dissimilarity measure $D(t_1, t_2)$ between the trajectories $\tau_1$ and $\tau_2$ is  defined as follows \cite{Liu2013}:
\begin{equation}
\label{eq:traj_dissim}
    D(\tau_1, \tau_2) = \frac{1}{n}\sum_{i=1}^n \lVert c_1^i-c_2^i\rVert_2
\end{equation}
\noindent where $n$ is a predefined constant denoting the number of trajectory segments chosen when normalizing.

The reward function is then modified as follows
\begin{equation}
\label{eq:traj_dissim_reward}
R_{TD}\left(s\right) = \left\{
        \begin{array}{ll}
            \frac{\gamma}{\mu}\sum_{i=1}^{\mu}D(t_s,t_i) &  s \in E \\[7pt]
            -\alpha - \beta \mathcal{D}(r_v, r_p) &  s \notin E, t\geq T \\[7pt]
            -\mathcal{M}(a \mid s) &  s \notin E, t < T
        \end{array}
    \right.
\end{equation}

\noindent where $\gamma$ is a user tunable parameter that controls the extent with which the solver is rewarded for discovering highly diverse failure scenarios. We define $\mu$ as min$(k,k')$ where $k$ is a user provided constant that specifies the number of top failure trajectories (ordered by total reward) that AST is asked to return and $k'$ is the number of failure trajectories that AST has already found. $\mathcal{D}(r_v, r_p)$ and $\mathcal{M}(a \mid s)$ remain unchanged from the generic AST reward function.

\section{Experiments}
\label{sec:experiments}
This section describes the experiments of the RSS and dissimilarity reward augmentation on the AV scenario. The software setup used to run the experiments is based on the Adaptive Stress Testing Toolbox\footnote{\url{https://ast-toolbox.readthedocs.io/en/latest/}}.

\subsection{RSS Reward Experiments}

The RSS reward augmentation was applied to the AV scenario with a single pedestrian and solved using the TRPO algorithm \cite{schulman2015trust}. TRPO trains a control policy for the environment that produces events with high likelihood of collision. The policy is then sampled for \num{1000} action sequences and those that ended with a collision between the car and the pedestrian were recorded. Action sequences were recorded for the generic AST algorithm and the RSS-augmented version. The RSS parameters shown in \cref{rss_params} were chosen to approximate plausible accelerations of an autonomous vehicle and pedestrian. A response time of $\rho=0$ was chosen because the SUT simulation has no in-built delay. For industrial applications, these parameters should be chosen with care, as they will affect the types of failures discovered by AST.

\begin{table}
\caption{RSS Parameters with $g = \SI{9.8}{\meter\per\second\squared}$ and $\rho=0$}
\vspace{-0.35cm}
\label{rss_params}
\begin{center}
\begin{small}
\begin{tabular}{@{}lrrr@{}}
\toprule
 & \hspace{0.25cm}$a_{\rm max}^{\rm acc}$\hspace{0.25cm} & \hspace{0.25cm}$a_{\rm min}^{\rm brk}$\hspace{0.25cm} & \hspace{0.25cm}$a_{\rm max}^{\rm brk}$\hspace{0.25cm}\\
\midrule
Lateral & \hspace{0.25cm}$0.1g$\hspace{0.25cm} & \hspace{0.25cm}$0.05g$\hspace{0.25cm} & \hspace{0.25cm}$-$\hspace{0.25cm}\\
Longitudinal & \hspace{0.25cm}$0.1g$\hspace{0.25cm}  & \hspace{0.25cm}$0.7g$\hspace{0.25cm} & \hspace{0.25cm}$0.7g$\hspace{0.25cm}\\
\bottomrule
\end{tabular}
\end{small}
\end{center}
\vskip -0.1in
\end{table}

\begin{figure}[ht]
    \centering
    \includegraphics[width=0.8\columnwidth]{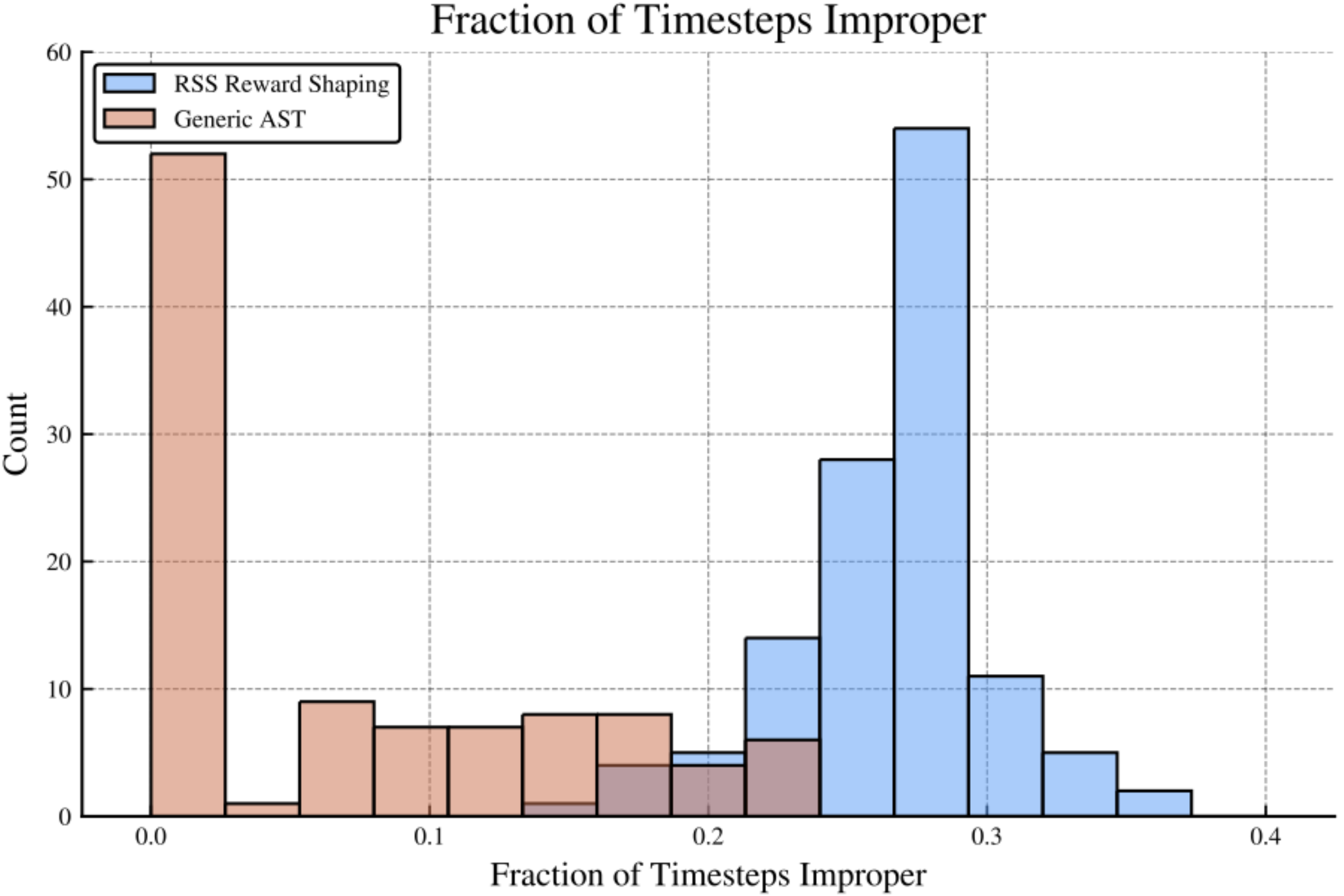}
    \caption{Distribution of trajectories over fraction of improper response for the generic AST reward function and the RSS reward function.}
    \label{fig:rss_response}
\end{figure}

\Cref{fig:rss_response} shows the aggregate results of the two trials. Without the RSS reward augmentation, the most common type of failure involves vehicle trajectories where the vehicle is not at fault, and a much smaller number of trajectories are found where the AV policy is to blame. For the RSS reward function, all trajectories have a non-zero fraction of improper timesteps, and in most trajectories, the AV policy behaved improperly for more than 25\% of the simulation timesteps.

\begin{figure}[ht]
    \centering
    \includegraphics[width=\columnwidth]{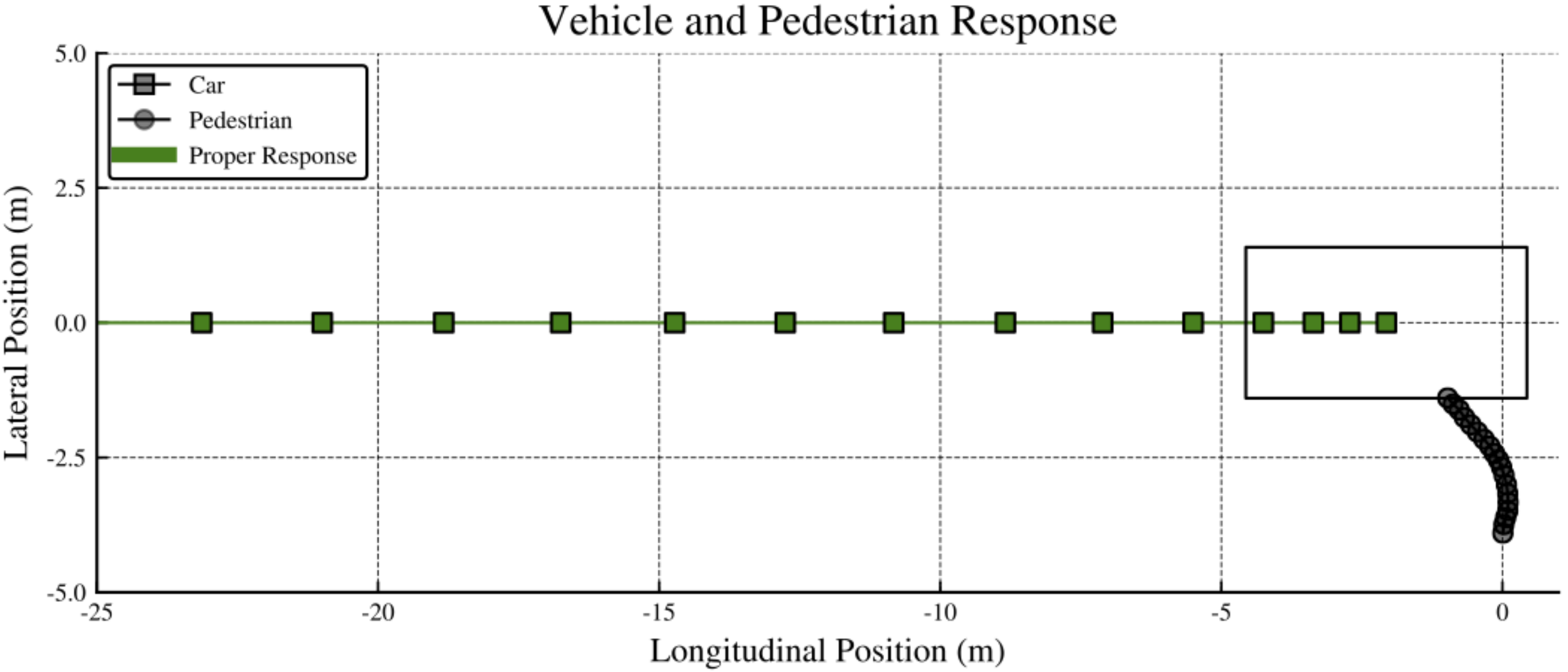}
    \caption{Example trajectory with the generic AST reward.}
    \label{fig:rss_response_generic}
    \vspace{-0.3cm}
\end{figure}

\begin{figure}[ht]
    \centering
    \includegraphics[width=\columnwidth]{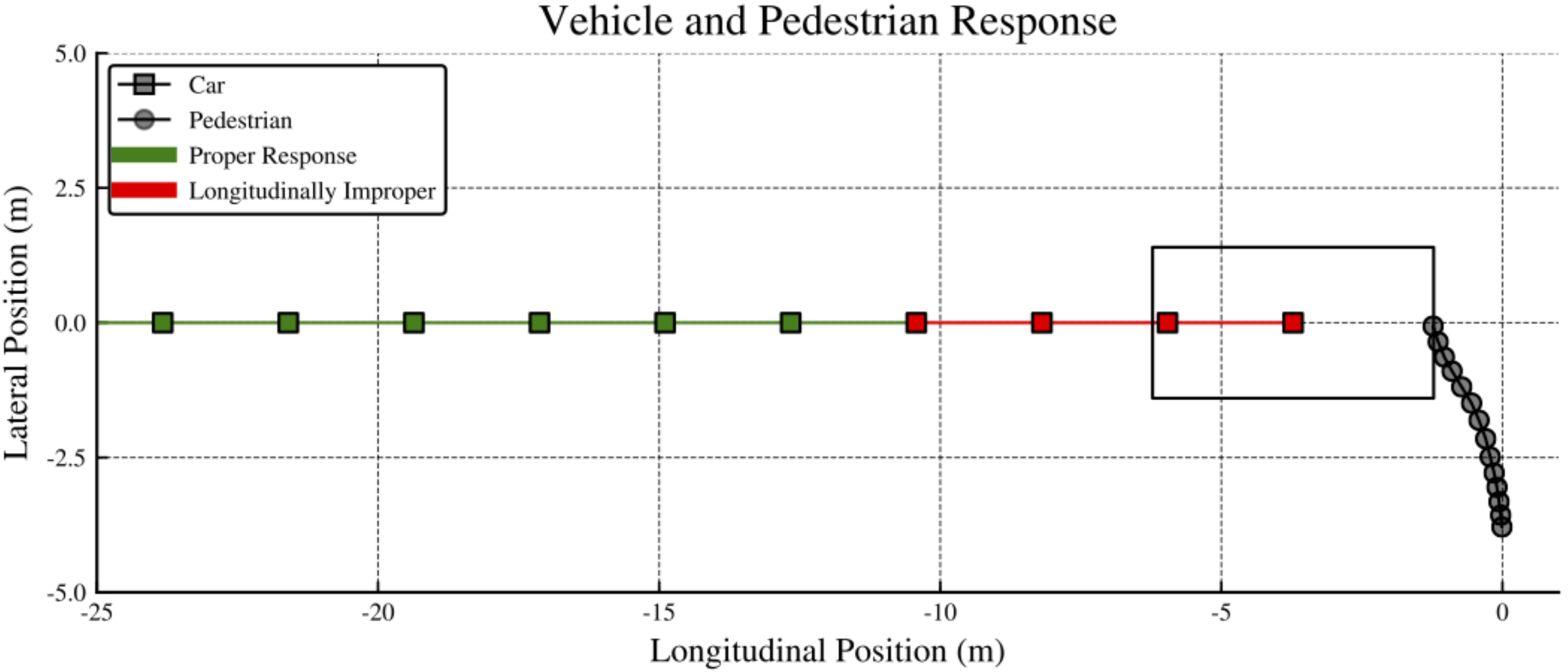}
    \caption{Example trajectory with RSS reward augmentation.}
    \label{fig:rss_response_rss_shape}
\end{figure}

Two representative trajectories were selected from two policies. \cref{fig:rss_response_generic} shows a trajectory from the generic AST trial where the pedestrian collides with the side of the car as it drives by. The car slows down as the pedestrian approaches, but because the pedestrian is encroaching laterally, the vehicle behaves properly by maintaining a zero lateral velocity. The trajectory from the RSS-augmented trial is shown in \cref{fig:rss_response_rss_shape} which shows that the vehicle entirely fails to stop when the pedestrian walks out in front of it. This failure is due to sensor noise and results in a much more relevant failure mode than the previous example. The trajectory shows that the vehicle behaved improperly in the last quarter of its trajectory and is clearly to blame for this collision.

\subsection{TD Reward Experiments}

The following experiments are conducted using a two vehicle and two pedestrian test case adapted from the scenario discussed in Section \ref{ast}. The vehicles drive on a single lane road and approach a crosswalk. One pedestrian is placed on either side of the crosswalk and they attempt to cross the road. The goal of the vehicles is to avoid coming too close to either pedestrian and each other. The set of failure states $E$ is set include all states $s$ where:
\begin{enumerate}
    \item Longitudinal ($x$) or lateral ($y$) positions between the lead car and any pedestrian is less than 0.5 meters.
    \item Longitudinal ($x$) position between the two cars is less than 0.5 meters.
\end{enumerate}

\noindent Initial conditions are shown in \cref{inital_conditions}.

\begin{table}[!h]
\caption{Starting state of each pedestrian and vehicle}
\vspace{-0.45cm}
\label{inital_conditions}
\begin{center}
\begin{small}
\begin{tabular}{@{}lrrrr@{}}
\toprule
 & $v_x$ (\si{\meter\per\second}) & $v_y$ (\si{\meter\per\second}) & $x$ (\si{\meter}) & $y$ (\si{\meter})\\
\midrule
Pedestrian 1 & 0 & 0.5 & 0 & \num{-3}\\
Pedestrian 2 & 0 & \num{-0.5} & 0 & 3\\
Car 1 & 11.1 & 0 & \num{-20} & 0\\
Car 2 & 12.5 & 0 & \num{-37} & 0\\
\bottomrule
\end{tabular}
\end{small}
\end{center}
\vskip -0.1in
\end{table}

\begin{figure}[!h]
    \vspace{-0.4cm}
    \centering
    \includegraphics[width=0.9\columnwidth]{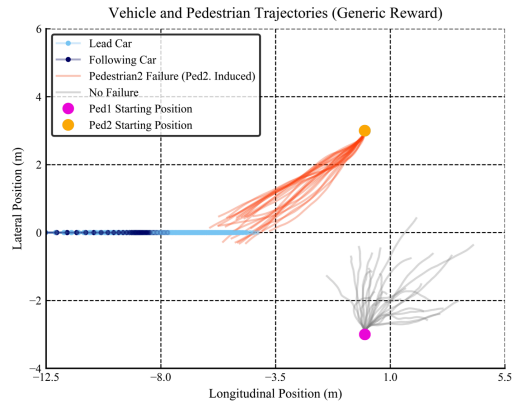}
    \caption{Trajectories returned with generic AST reward (two pedestrian/two car scenario).}
    \label{fig:generic_reward_all_traj}
     \vspace{-0.2cm}
\end{figure}

\Cref{fig:generic_reward_all_traj} shows all failure trajectories found after running AST without reward augmentation using an MCTS solver. In this setup, the results show a strong bias towards a particular scenario where the lead vehicle stops well ahead of the crosswalk and a pedestrian then moves too close to the vehicle, causing the failure (pedestrian induced). \cref{failure_type_summary} shows the number of various failure types found using AST. We see that without reward augmentation, AST discovers and returns a single type of failure. The results demonstrate the drawbacks of using this baseline version of AST for failure detection in autonomous vehicle systems. While the method is able to identify the maximum required number of failures (25 in this case), the results lack diversity and converge to one failure mode. Furthermore, this mode lacks relevance as it identifies a scenario where the pedestrian runs towards a stopped vehicle; a situation that is unavoidable by the car. 

\begin{table}
\caption{Number of failures obtained using AST}
\vspace{-0.5cm}
\label{failure_type_summary}
\begin{center}
\begin{small}
\begin{tabular}{@{}m{2.8cm}  rr@{}}
\toprule
Failure Type & Generic Reward & TD Reward \\
\midrule
Vehicle/Pedestrian (Vehicle Induced) & \hspace{1.2cm}0 & \hspace{0.8cm}4 \\
\midrule
Vehicle/Pedestrian (Ped. Induced) & \hspace{1.1cm}25 & \hspace{0.7cm}15\\
\midrule
Vehicle/Vehicle & \hspace{1.2cm}0 & \hspace{0.8cm}6\\
\bottomrule
\end{tabular}
\end{small}
\end{center}
\vspace{-0.65cm}
\end{table}

\begin{figure}[!ht]
    \vspace{-0.4cm}
    \centering
    \includegraphics[width=0.9\columnwidth]{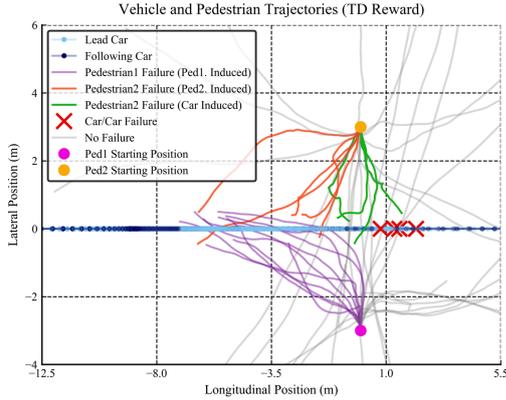}
    \caption{Trajectories returned with TD reward augmentation (two pedestrian/two car scenario).}
    \label{fig:td_reward_all_traj}
    \vspace{-0.4cm}
\end{figure}

\Cref{fig:td_reward_all_traj} shows failure trajectories found using AST with trajectory dissimilarity reward augmentation and an MCTS solver that is run for the same number of iterations as the above case. By incorporating the dissimilarity metric, the solver is able to explore a larger subset of the failure space. \Cref{failure_type_summary} shows that the trajectories obtained now consist of other failure modes in addition to the pedestrian induced ones discovered using the generic AST reward function. The green trajectories in \cref{fig:td_reward_all_traj} indicate cases where the lead car fails to apply adequate braking and approaches too close to a pedestrian within the crosswalk region (vehicle induced vehicle/pedestrian failure), while the red crosses indicate failure cases where the second vehicle follows too closely with the lead vehicle (vehicle/vehicle failure). Both cases may indicate possible vehicle policy shortcomings and show that when compared with generic AST, trajectory dissimilarity reward augmentation allows for the discovery of relevant weaknesses of this AV policy.


\section{CONCLUSIONS}
\label{sec:conclusions}
This paper proposed two reward augmentation methods that allow Adaptive Stress Testing to be more suitable for autonomous vehicle validation. We showed the limitations of the current AST method in finding relevant failure cases and its tendency to repeatedly converge on single failures. By augmenting the reward function in AST with domain relevant information from RSS and trajectory dissimilarity, we showed that we were able to discover a more diverse set of failures. In the case of RSS rewards, the AST method found trajectories with a high percentage of improper vehicle responses, while with trajectory dissimilarity rewards, AST found trajectories with high failure diversity. Our results indicate that with reward augmentation, AST is able to find more useful failures that can aid in the validation of autonomous vehicles. With these modifications, we suggest that AST can be used as an efficient simulation based tool to find relevant weaknesses in an AV policy. Future work will involve incorporating these two reward augmentation methods into a single reward framework to discover a diverse set of failures with improper vehicle responses.

\addtolength{\textheight}{-12cm}   



\printbibliography



\end{document}